\documentclass[lettersize,journal]{IEEEtran}
\pdfoutput=1
\usepackage{amsmath,amsfonts}
\usepackage{algorithmic}
\usepackage{algorithm}
\usepackage{array}
\usepackage[caption=false,font=normalsize,labelfont=sf,textfont=sf]{subfig}
\usepackage{textcomp}
\usepackage{stfloats}
\usepackage{url}
\usepackage{verbatim}
\usepackage{graphicx}
\usepackage{}
\usepackage{cite}
\usepackage{booktabs}
\usepackage{adjustbox}
\usepackage{multirow}
\usepackage{indentfirst}
\usepackage{orcidlink}
\usepackage{CJKutf8}

\begin{document}
\begin{CJK}{UTF8}{gbsn}
\title{Efficient Remote Sensing Segmentation With Generative Adversarial Transformer}

\author{Luyi~Qiu \orcidlink{0000-0002-8158-4604},
        Dayu~Yu \orcidlink{0000-0003-1720-8302}, 
        Xiaofeng Zhang \orcidlink{0000-0002-7185-4682}
        and~Chenxiao~Zhang
\thanks{Manuscript received April xx, xxxx; revised September xx, xxxx.	The work was supported by National Natural Science Foundation of China (No.42201396), China National Postdoctoral Program for Innovative Talents (No.BX2021223), Natural Science Foundation of Hubei Province (No.2022CFB668), and China Postdoctoral Science Foundation (No.2021M702510).}
\thanks{Luyi Qiu is from School of Information and Software Engineering, University of Electronic Science and Technology of China, Chengdu, China 610054. E-mail: 202021090124@std.uestc.edu.cn.}
\thanks{Dayu Yu and Chenxiao Zhang are from School of Remote Sensing and Information Engineering, Wuhan University, Wuhan, China 430079. Correspondence: dayuyu@whu.edu.cn. }
\thanks{Xiaofeng Zhang is from School of Electronic Information and Electrical Engineering, Shanghai Jiao Tong University, Shanghai, China 200240. E-mail: framebreak@sjtu.edu.cn}

}

\markboth{Journal of \LaTeX\ Class Files,~Vol.~13, No.~9, September~2022}%
{Qiu \MakeLowercase{\textit{et al.}}: Bare Demo of IEEEtran.cls for Journals}
\maketitle

\begin{abstract}
Most deep learning methods that achieve high segmentation accuracy require deep network architectures that are too heavy and complex to run on embedded devices with limited storage and memory space. To address this issue, this paper proposes an efficient Generative Adversarial Transfomer (GATrans) for achieving high-precision semantic segmentation while maintaining an extremely efficient size. The framework utilizes a Global Transformer Network (GTNet) as the generator, efficiently extracting multi-level features through residual connections. GTNet employs global transformer blocks with progressively linear computational complexity to reassign global features based on a learnable similarity function. To focus on object-level and pixel-level information, the GATrans optimizes the objective function by combining structural similarity losses. We validate the effectiveness of our approach through extensive experiments on the Vaihingen dataset, achieving an average F1 score of 90.17\% and an overall accuracy of 91.92\%.
\end{abstract}
  

\section{Introduction}
\label{sec:1}

\IEEEPARstart{S}{emantic} segmentation, as a significant task in image processing, has found application in various practical scenarios such as autonomous driving, precision agriculture, and urban analysis \cite{1}. Over the past decade, inspired by the success of deep learning in high-level visual tasks, a considerable amount of work has been devoted to using deep convolutional neural networks (DCNNs) for semantic segmentation of remote sensing images \cite{12,13,14}. The inherent characteristics of geographical objects in remote sensing images, including their multi-scale nature, random appearances, and varied locations, pose a challenging problem for DCNNs. Furthermore, many existing DCNN methods have a large number of parameters and require significant computational resources, making it difficult to run them on devices with limited memory capacity.

In contrast to the independent predictions made by DCNNs, generative adversarial networks (GANs) \cite{7} applied to dense prediction tasks treat the segmentation model as a generator and optimize the weights of the generator through a generative-adversarial strategy, without increasing the number of parameters, enhancing the spatial contiguity of predictions \cite{6}. Consequently, several studies aim to explore the contribution of the generative adversarial strategy to image processing \cite{2,5}. However, accomplishing image segmentation through the generative adversarial strategy comes with certain flaws. Luc et al \cite{6}. pointed out that the fake/real scalar of the adversarial loss alone lacks sufficient gradients to stabilize the training framework.

Meanwhile, very high-resolution (VHR) images contain multi-scale details of objects and suffer from class imbalance issues \cite{13}. Some efforts have been made to improve the recognition ability through enhancing multi-scale fusion modules \cite{12} and architectures \cite{3}. However, these methods only implicitly capture global relationships through repeated convolutional operations, lacking the ability to establish dependencies among features and fully utilize global contextual information. In contrast, Transformer, since its introduction to the field of computer vision, has quickly become a research hotspot due to its capability to learn explicit global and long-range semantic features \cite{4,8}. Nevertheless, previous studies have overlooked the non-local textures with low similarity, which might offer richer detail information than highly similar features \cite{11}. Additionally, although global features can be captured, Transformer also result in higher computational complexity because each position's feature needs to be computed and interacted with other positions.
 
In this paper, we propose an efficient Generative Adversarial Transformer (GATrans) for achieving high-precision semantic segmentation of VHR images while maintaining an extremely efficient size. The framework adopts a Global Transformer Network (GTNet) to capture long-range contextual dependencies and optimizes the weights of the generator through a generative-adversarial strategy. The GATrans employs a global Transformer generator to capture long-range dependency features and focuses on object-level information by optimizing an objective function that combines structural similarity loss and adversarial loss. The main contributions of this paper are as follows:

\begin{enumerate}
	\item We propose an efficient GATrans framework for VHR image segmentation, which strengthens the spatial contiguity of predictions through a generative-adversarial strategy without increasing the number of parameters and achieves state-of-the-art performance.
	\item The efficient GTNet is proposed as a generator to extract multi-level features. It utilizes a global Transformer block with progressively linear computational complexity to reassign global features based on a learnable similarity function.
	\item Extensive experiments are conducted on the Vaihingen dataset to evaluate the performance of the GATrans framework, and the GATrans achieves better effectiveness than advanced methods.
\end{enumerate}

\section{Method}\label{sec3}
\subsection{Overall Architecture}

\begin{figure}[htb]
 \includegraphics[width=\linewidth]{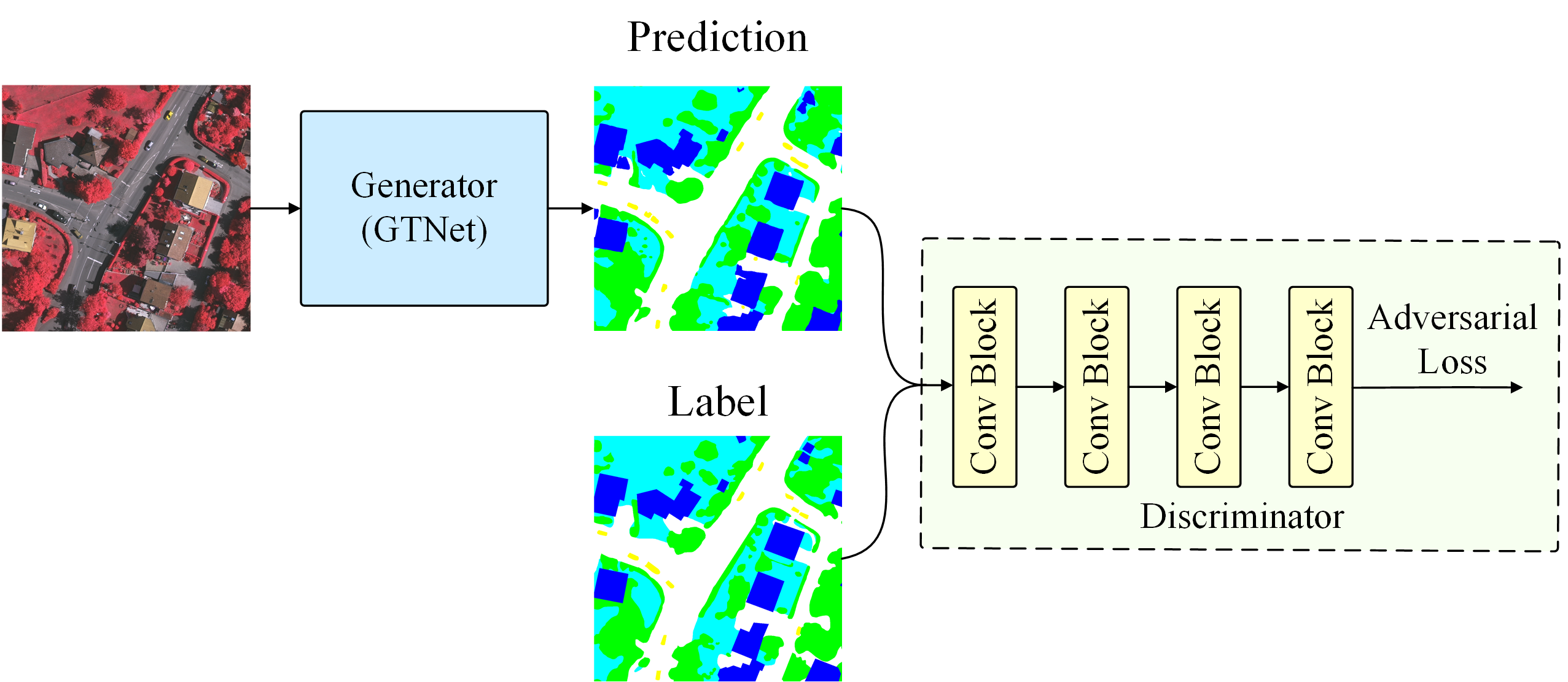}
	\caption{The overview of the Generative Adversarial Transformer (GATrans). \label{0fig}}
\end{figure}

As shown in figure \ref{0fig}, the GATrans framework ultizies the GTNet as a generator to synthesize predictions and confuse the discriminator. Then, the GATrans framework concatenates labels as conditioned auxiliary information with the predictions generated by the generator and inputs them into the discriminator. The discriminator, consisting of a 4-layer network, aims to distinguish between real and fake synthesized images. Additionally, the GTNet framework combines structural similarity loss with objective loss to increase the complexity of the gradient in the training process, making the framework could focus on pixel-level and object-level information.  

\subsection{Global Transformer Network} \label{sec3.2}


Within GTNet, the encoder incorporates a patch partition layer, which divides the image into non-overlapping patches of a fixed dimension. These patches are then inputted into residual blocks, global transformer blocks, and patch merging layers. The residual blocks and global transformer (GT) blocks capture image features, while the patch merging layer performs downsampling operations. Moreover, the decoder employs deconvolution to upsample image sizes and incorporates skip connections to fuse low and high-level features, as shown in figure \ref{fig2}.

\begin{figure}[htb]
	\centering
 \includegraphics[width=\linewidth]{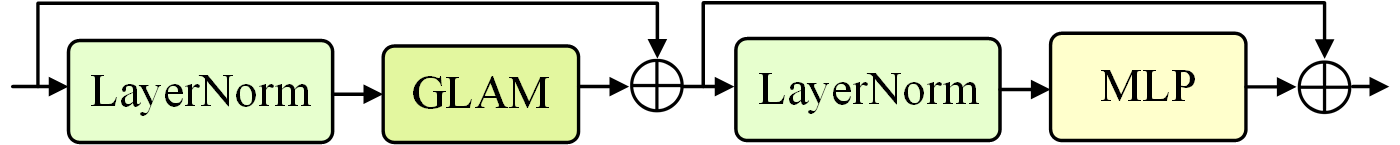}
	\caption{The overview of the global transformer block. \label{fig3}}
\end{figure}

\begin{figure}[htb]
 \includegraphics[width=\linewidth]{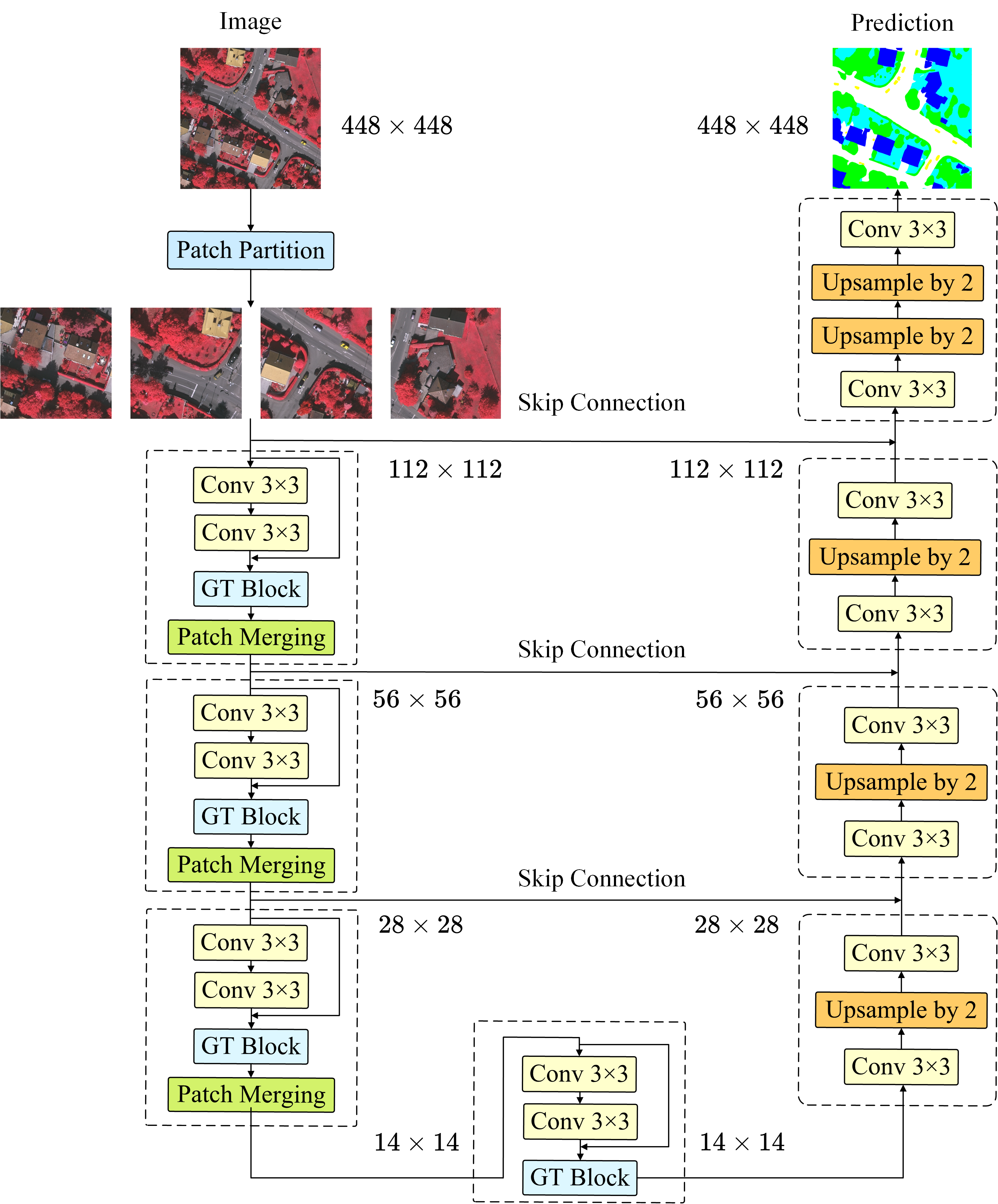}
	\caption{The overview of the Global Transformer Network (GTNet). \label{fig2}}
\end{figure}

The global transformer block, depicted in Figure \ref{fig3}, consists of layer normalization layers, a multi-layer perceptron with a GELU activation function, and residual connections. Additionally, the global learnable attention module (GLAM) plays a crucial role within the global transformer block, allowing for the exploration of global information and enhancing the accuracy of image segmentation, particularly when dealing with complex objects.



\begin{figure}[htb]
	\centering
	\includegraphics[width=0.85\linewidth]{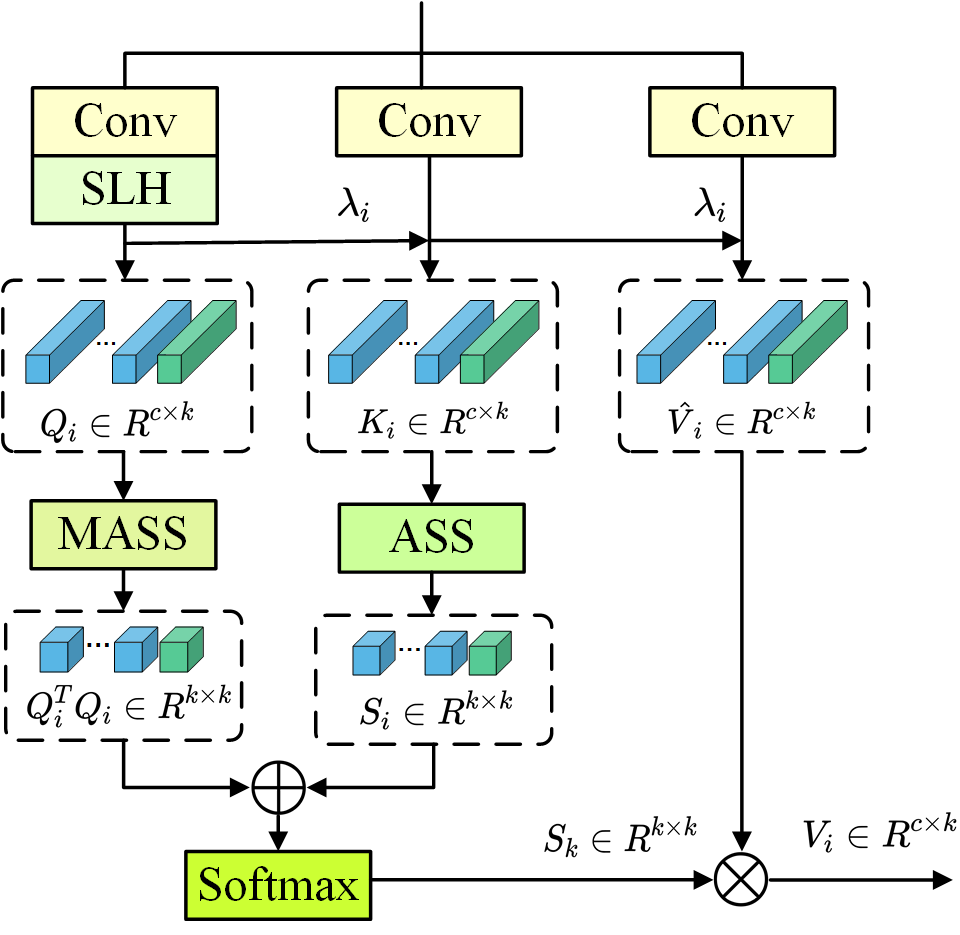}
	\caption{The overview of the GLAM module. \label{fig4}}
\end{figure}

As shown in figure \ref{fig4}, the GLAM modules capture global information by aggregating similar features from the input features $X \in \mathbb{R}^{h \times w \times c}$, which are subsequently reshaped to a dimensional representation $X' \in \mathbb{R}^{h \times w \times c}$. The calculation process of the query $x_i$ is illustrated by Equation \ref{eq3}.
\begin{equation}
\begin{aligned}
\label{eq3}
f(x_i)= \sum_{x_j\in\lambda _i} \frac{exp(s(x_i,x_j))}{\sum_{x_k\in\lambda _i}exp(s(x_i,x_k)}\phi _v(x_j)
\end{aligned}
\end{equation}
where $n=hw$, $x_i$ represents the $i$-th vector in $X'$. The function $\phi_v(\cdot)$ is utilized to generate value vectors, $\lambda_i$ denotes the features assigned to one query bucket using the super-bit locality-sensitive hashing (SLH) algorithm \cite{10}, and $s(\cdot,\cdot)$ measures the similarity between vectors.

Firstly, the GLAM module utilizes the SLH method to hash global features into query buckets, effectively reducing computational complexity. The SLH algorithm estimates similarity to ensure that similar features are more likely to be assigned to the same hash bucket. Thus, the SLH algorithm performs an appropriate preprocessing step for the GLAM module. As shown in Equation \ref{eq7}, when the global features have $d$ buckets and the query has a dimension of $c$, the SLH algorithm projects the query onto an orthonormal matrix $M \in \mathbb{R}^{b \times c}$.

\begin{equation}
\begin{aligned}
\label{eq7}
x_{i}^{'}=Mx_{i}
\end{aligned}
\end{equation}

Then, the SLH assigns the hash bucket of $x_{i}$ as $h(x_{i}^{'})=\text{argmax}(x_{i}^{'})$, where $\text{argmax}(\cdot)$ finds the index of the maximum value from $x_{i}^{'}$. As shown in Equation \ref{eq8}, global features are hashed into the same bucket $\lambda _{i}$ as the query $x_{i}$.
\begin{equation}
\begin{aligned}
\label{eq8}
\lambda _{i}=\left \{ x_j|hash(x_{i}^{'}) =hash(x_{j}^{'}) \right \}
\end{aligned}
\end{equation}

The SLH performs batch matrix multiplication for all queries, which helps the GLAM module reduce computational complexity.

Inspired by the similarity function proposed in \cite{9}, the GLAM module adopts a hidden layer network (FNN) as an adaptive similarity function. This network consists of an adaptive similarity function (ASS) $s_{ASS}(\cdot)$ and a fixed similarity function (MASS) $s_{MASS}(\cdot,\cdot)$, as shown in Equation \ref{eq4}.

\begin{equation}
\begin{aligned}
\label{eq4}
 s(x_{i},x_{j})=s_{ASS}^{j}(x_{i})+s_{MASS}(x_{i},x_{j})
\end{aligned}
\end{equation}
where $s_{ASS}^{j}(x_{i})$ indicates the $j-th$ GLAM module.

The ASS similarity function adaptively adjusts similarity scores through two learnable convolutions, as shown in Equation \ref{eq6}.
\begin{equation}
\begin{aligned}
\label{eq6}
s_{ASS}(x_i)=W_2\cdot ReLU(W_1\phi_l(x_i)+b_1)+b_2
\end{aligned}
\end{equation}
where $ReLU(\cdot)$ is a activation function, $W_1, W_2 \in R^{n\times c} $, $b_1, b_2\in R^{n}$.

And the MASS similarity function involves the dot product operation, as shown in Equation \ref{eq5}.
\begin{equation}
\begin{aligned}
\label{eq5}
 s_{MASS}(x_{i},x_{j})=\phi_{q}(x_{i})^T\phi_{k}(x_{j})
\end{aligned}
\end{equation}
where $\phi_{q}(\cdot)$ and $\phi_{k}(\cdot)$ are used to generate query and key through vector transformation.

\subsection{Loss Function} \label{sec3.4}
In the training process, the generator $G$ aims to obtain the optimal discriminator $D_G^*$ by maximizing the objective function $V(D, G)$. This maximization enhances the discriminator's ability to distinguish between real scene images and images generated by the generator. Mathematically, we can express it as $D_G^*=\arg \left(\max _D V(D, G)\right)$. Conversely, the discriminator $D$ aims to obtain the optimal generator $G_D^*$ by minimizing the same objective function, denoted as $D_G^*=\arg \left(\min _G V(D, G)\right)$. The GAN achieves the optimal generator $G_D^*$ when the distribution of the generated images is equal to the distribution of real images. In Equation \ref{eq14}, the input of the generator is represented by $x$, and the label is indicated by $y$.

\begin{equation}
\begin{aligned}
\label{eq14}
\min _G \max _{{D}} {V}({D}, {G})=&{E}_{{y} \sim {p}_{\text {data}}({y})}[\log D({y})]\\
&+{E}_{{x} \sim {p}_{{x}}({x})}[\log (1-{D}({G}({x})))] 
\end{aligned}
\end{equation}

In the GATrans, the generative loss is implemented by the cross-entropy loss. Additionally, the adversarial loss is defined as shown in Equation \ref{eq9}. Given an input $x$, a label $y$, $G(x)$ represents the output of the generator, and $D(\cdot)$ represents the output of the discriminator. The term $l_{MSE}(y, G(x))$ calculates the pixel-level distance between the label and the prediction generated by the generator. On the other hand, $l_{Dice}(y, G(x))$ evaluates the region-level differences between the label and the generated prediction. The parameter $\alpha$ and $\mu$ are set at 0.5, indicating an equal weighting between the loss terms. Consequently, the structural similarity loss contributes to reducing both pixel-level and object-level differences between the label and the generated prediction, aiming to improve the overall performance of the framework.

\begin{equation}
\begin{aligned}
\label{eq9}
Loss_D(G, D)=&-E_{x, y}(\log (D(y)) + \log (1-D(G(x))))\\
&+\mu  \cdot l_{M S E}(y, G(x)) \\&
+\alpha \cdot l_{\text{Dice}}(y, G(x))
\end{aligned}
\end{equation}

\section{Experiments}\label{sec4}
\subsection{Experimental Settings}
\subsubsection{Dataset}

The 33 IRRG images with approximately 2494 \texttimes{} 2064 pixels from the Vaihingen dataset is selected as the experimental dataset, which encompasses five categories (as illustrated in Figure 5). In the experiments, 16 images are assigned to the training set, 17 to the test set, and 2 to the validation set.

\begin{figure}[htb]
	\centering
	\includegraphics[width=0.9\linewidth]{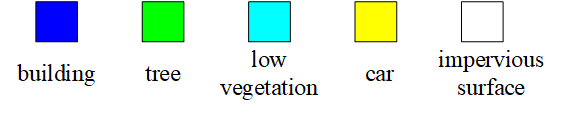}
	\caption{The five categories of the Vaihingen dataset. \label{data}}
\end{figure}

\subsubsection{Evaluation Metrics}
The evaluation of the GATrans framework utilizes classical metrics such as the F1 score and overall accuracy (OA) to assess its performance. As Equation \ref{eq10} and \ref{eq11}, where TP, TN, FP, and FN denote true positive, true negative, false positive, and false negative.
\begin{equation}
\begin{aligned}
\label{eq10}
F1 = \frac{2TP}{2TP+FP+FN} 
\end{aligned}
\end{equation}
\begin{equation}
\begin{aligned}
\label{eq11}
OA =\frac{TP+TN}{TP+FP+TN+FN} 
\end{aligned}
\end{equation}


\subsubsection{Implementation Details}
In the training phase, experiments use flip, random rotation, and size scale transformation to increase the number of images. The GATrans framework utilizes the Adam optimizer with a momentum of 0.9 and a weight decay setting of 0.0001, where parameters $\beta_{1}$ and $\beta_{2}$ are 0.9 and 0.99, and the initial learning rate is 0.001. Moreover, the GATrans framework adopts the slide-window method for images, where the input size is 448 $\times$ 448 pixels, the overlap stride is 32 pixels, and the batch size is 16.

\begin{table*}[htb]
	\centering

	\caption{Quantified results of ablation experiments on the test set.\label{tab1}}
	\begin{tabular}{@{}cccccccccccccc@{}}
		\toprule
		\multirow{2}{*}{Unet} & \multirow{2}{*}{ResUnet50} & \multirow{2}{*}{\begin{tabular}[c]{@{}c@{}}Attention   \\ Unet\end{tabular}} & \multirow{2}{*}{\begin{tabular}[c]{@{}c@{}}Swin   \\ Unet\end{tabular}} & \multirow{2}{*}{GTNet} & \multirow{2}{*}{GAN} & \multirow{2}{*}{\begin{tabular}[c]{@{}c@{}}Structural \\Similarity Loss\end{tabular}} & \multicolumn{5}{c}{F1   score}                  & \multirow{2}{*}{OA} & \multirow{2}{*}{\begin{tabular}[c]{@{}c@{}}Mean F1\end{tabular}} \\ \cmidrule(lr){8-12}
		&                            &                                                                              &                                                                         &                        &                      &                                                                               & Imp surf & Building & Low veg & Tree  & Car   &                     &                                                                     \\ \midrule
		\checkmark                     &                            &                                                                              &                                                                         &                        &                      &                                                                               & 91.26     & 93.88    & 82.47    & 88.79 & 84.16 & 90.19               & 88.112                                                              \\
		& \checkmark                          &                                                                              &                                                                         &                        &                      &                                                                               & 91.48     & 93.98    & 83.03    & 89.07 & 79.42 & 90.41               & 87.396                                                              \\
		&                            & \checkmark                                                                           &                                                                         &                        &                      &                                                                               & 92.34     & 94.74    & 83.25    & 88.41 & 86.68 & 90.57               & 89.084                                                              \\
		&                            &                                                                              &\checkmark                                                                       &                        &                      &                                                                               & 88.83     & 90.51    & 80.69    & 86.97 & 86.37 & 89.55               & 86.674                                                              \\
		&                            &                                                                              &                                                                         & \checkmark                      &                      &                                                                               & 93.07     & 96.19    & 83.59    & 89.39 & 86.60 & 91.67               & 89.768                                                              \\
		&                            &                                                                              &                                                                         & \checkmark                      &\checkmark                    &                                                                               & 93.25     & 96.09    & 84.55    & 89.62 & 86.11 & 91.87               & 89.924                                                              \\
		&                            &                                                                              &                                                                         & \checkmark                      & \checkmark                    & \checkmark                                                                             & 93.16     & 96.12    & 84.68    & 89.83 & 87.06 & 91.92               & 90.170                                                               \\ \bottomrule
	\end{tabular}
\end{table*}

\begin{figure*}[htbp]
	\centering
	\begin{minipage}{\linewidth}
		\centering
		{\includegraphics[width=.1\linewidth]{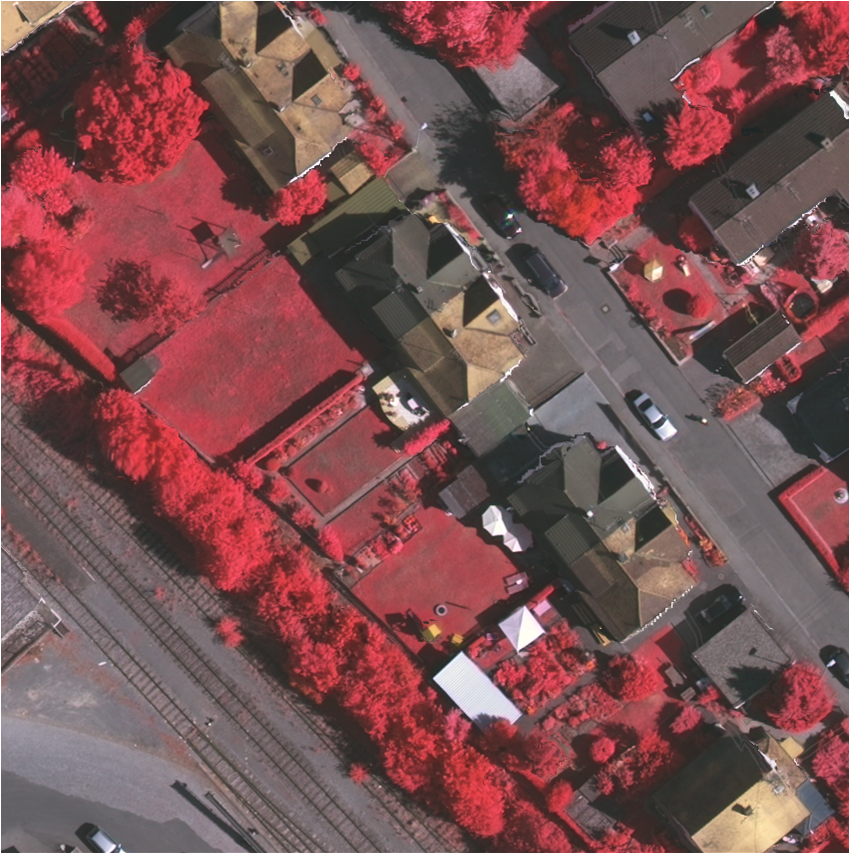}}
		{\includegraphics[width=.1\linewidth]{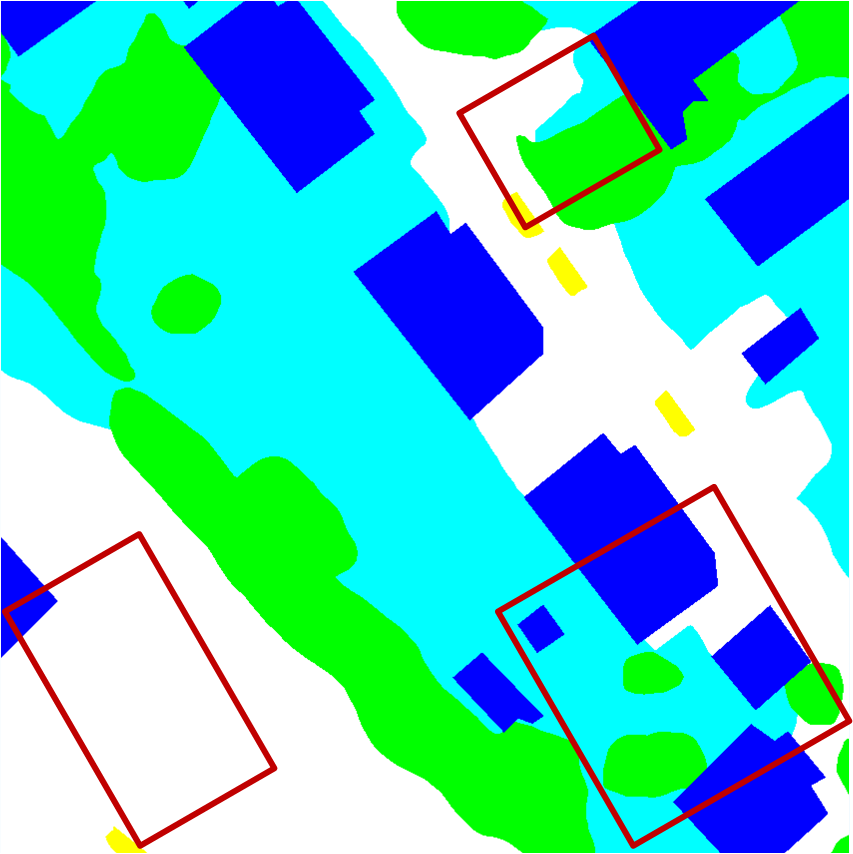}}
		{\includegraphics[width=.1\linewidth]{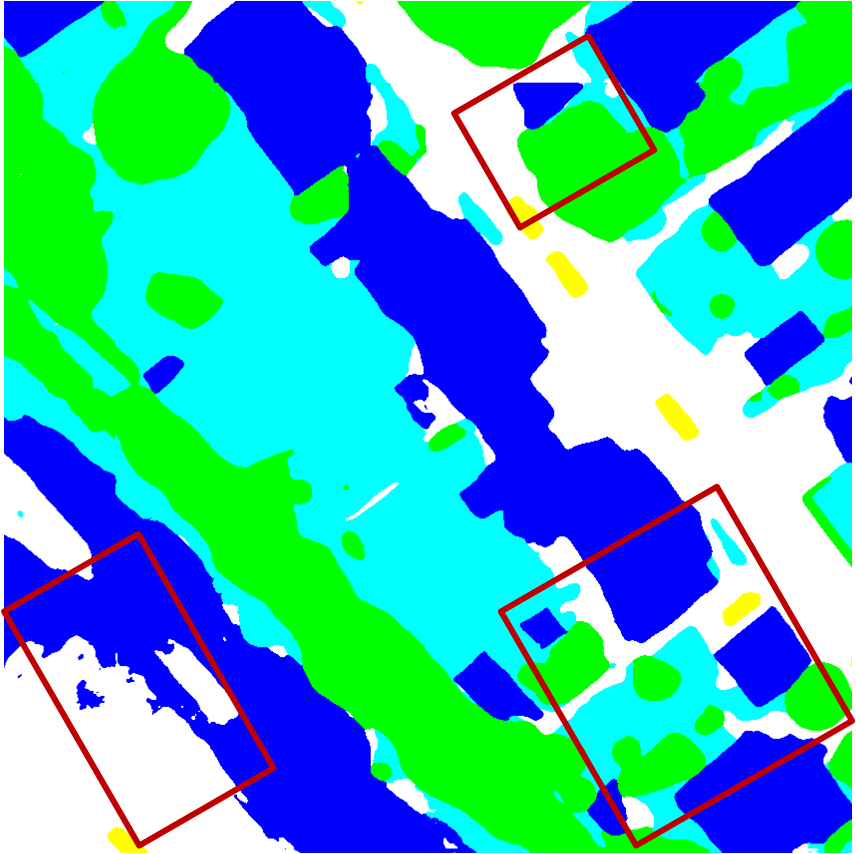}}
		{\includegraphics[width=.1\linewidth]{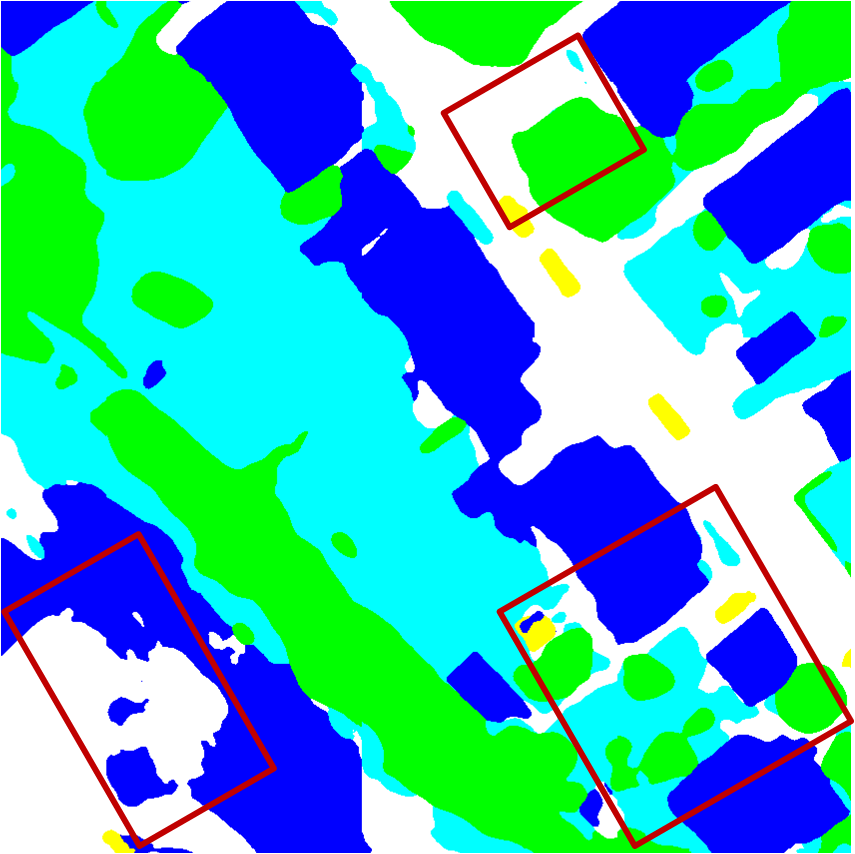}}
		{\includegraphics[width=.1\linewidth]{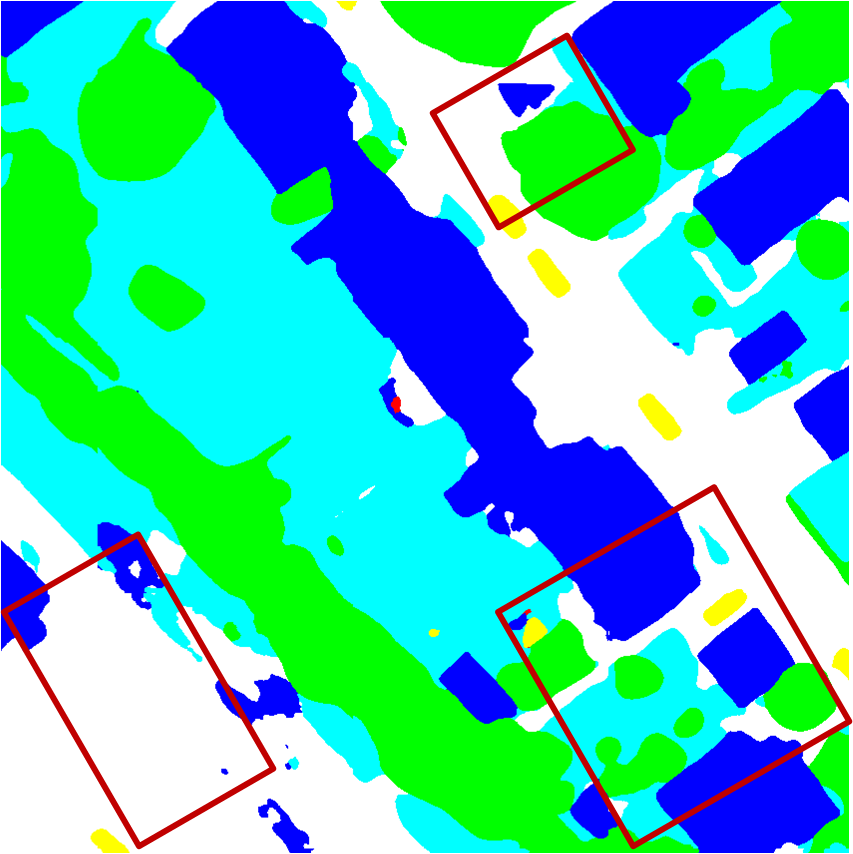}}
		{\includegraphics[width=.1\linewidth]{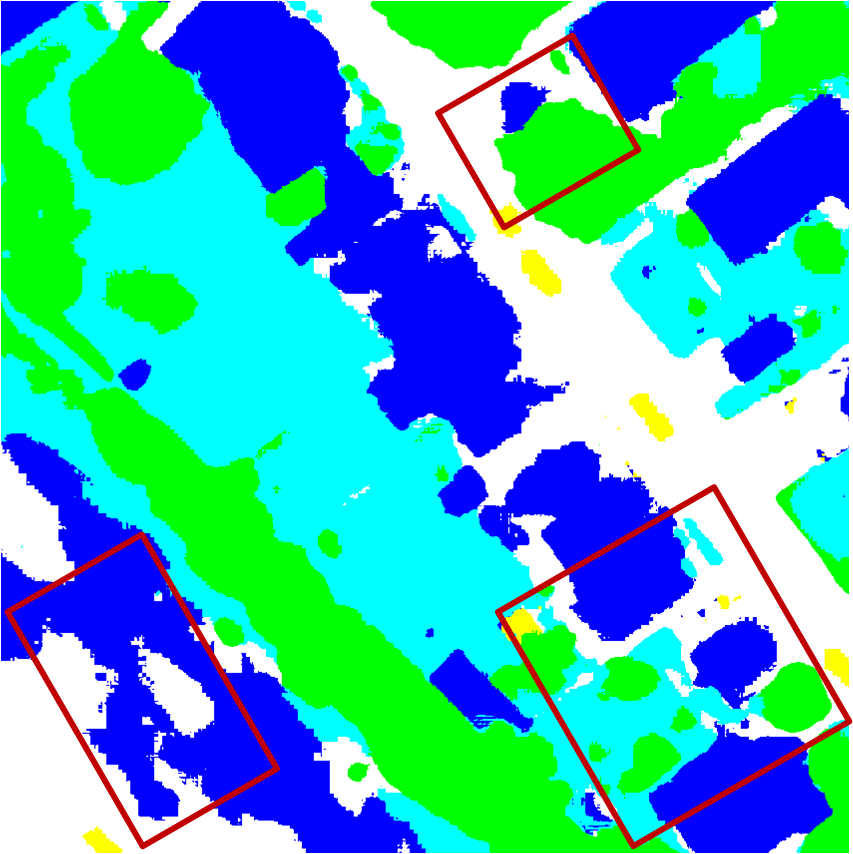}}
		{\includegraphics[width=.1\linewidth]{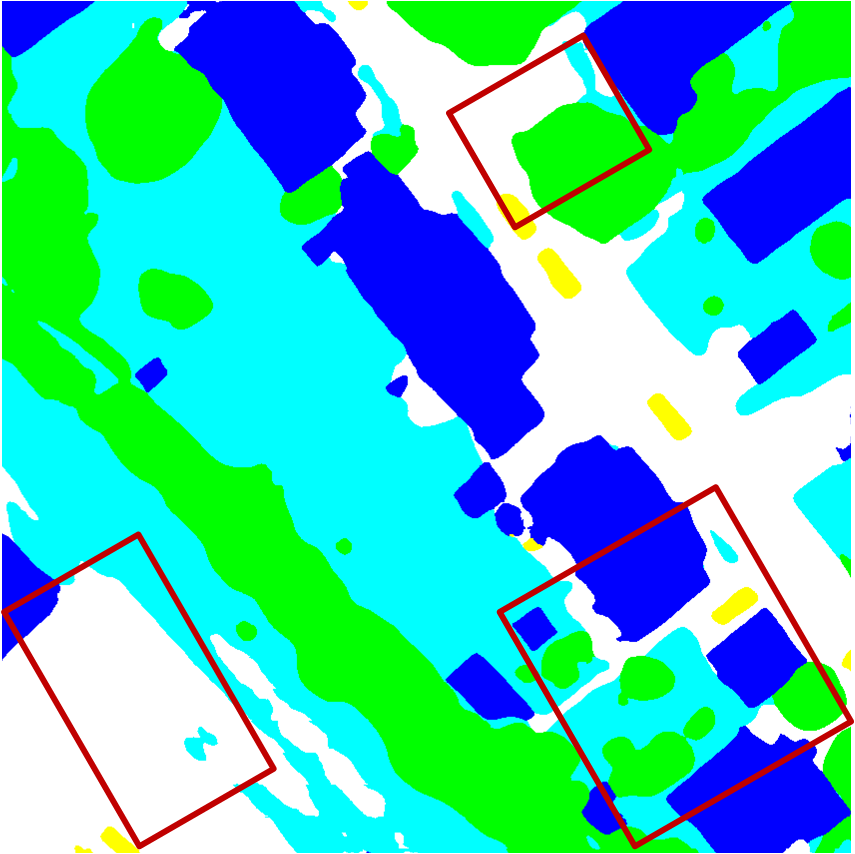}}
		{\includegraphics[width=.1\linewidth]{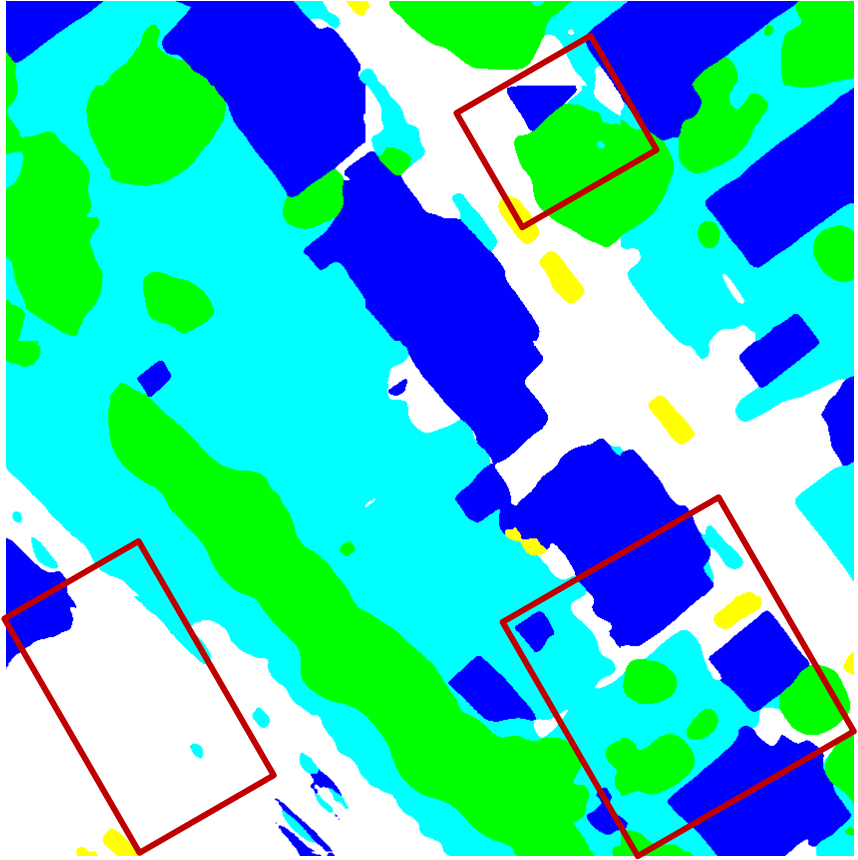}}
		{\includegraphics[width=.1\linewidth]{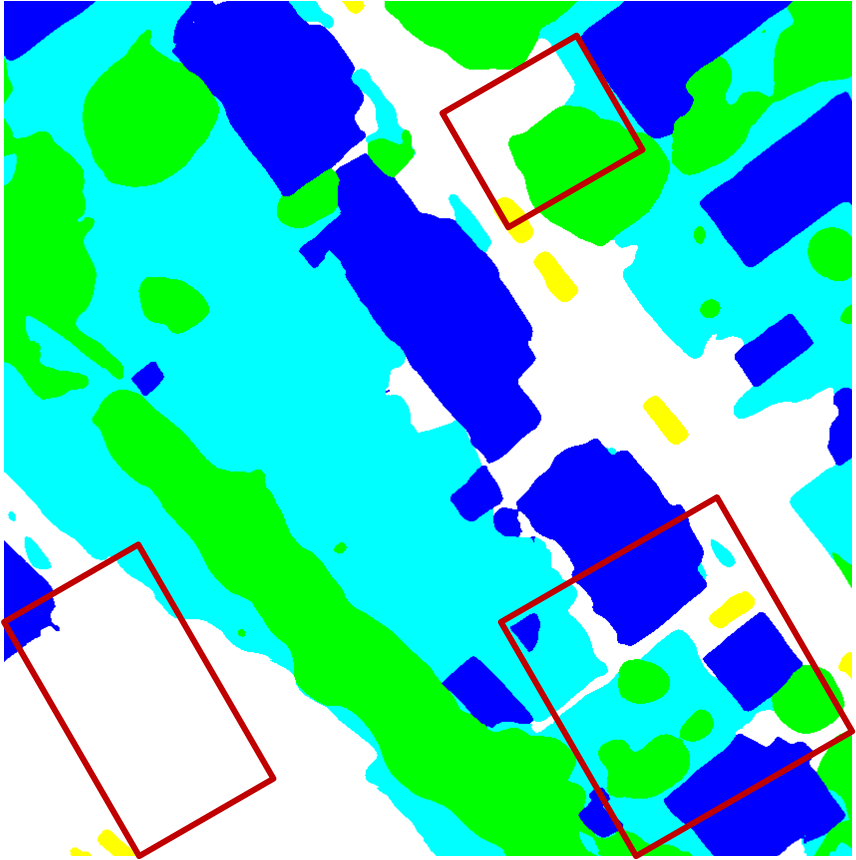}}
	\end{minipage}
	
	\vspace{0.1cm}
	
	\begin{minipage}{\linewidth}
		\centering
		{\includegraphics[width=.1\linewidth]{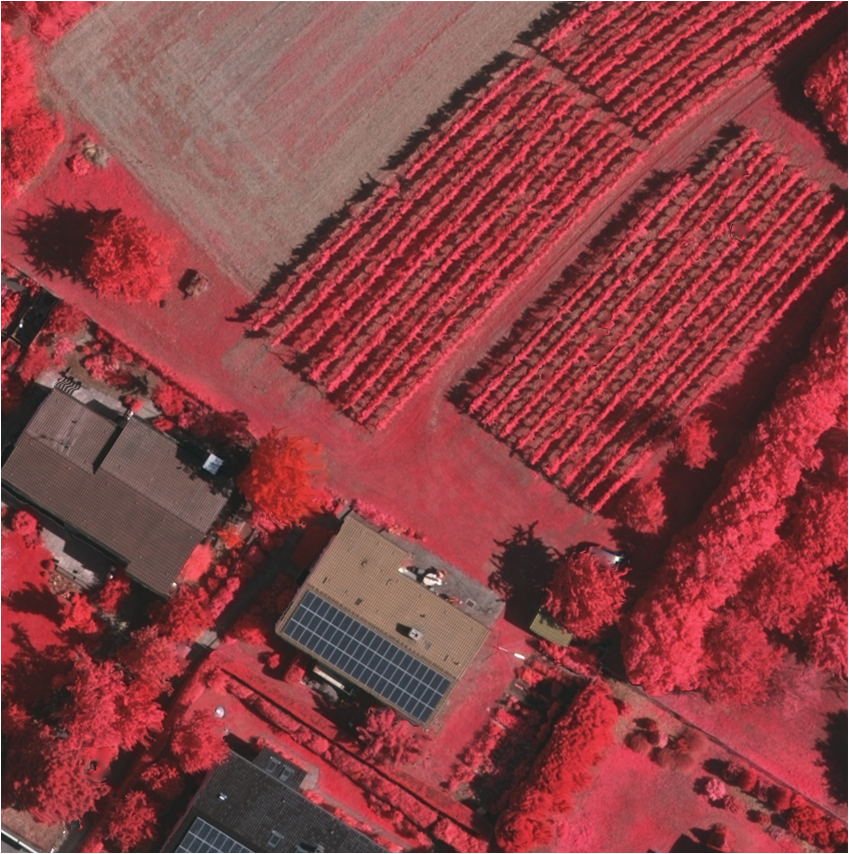}}
		{\includegraphics[width=.1\linewidth]{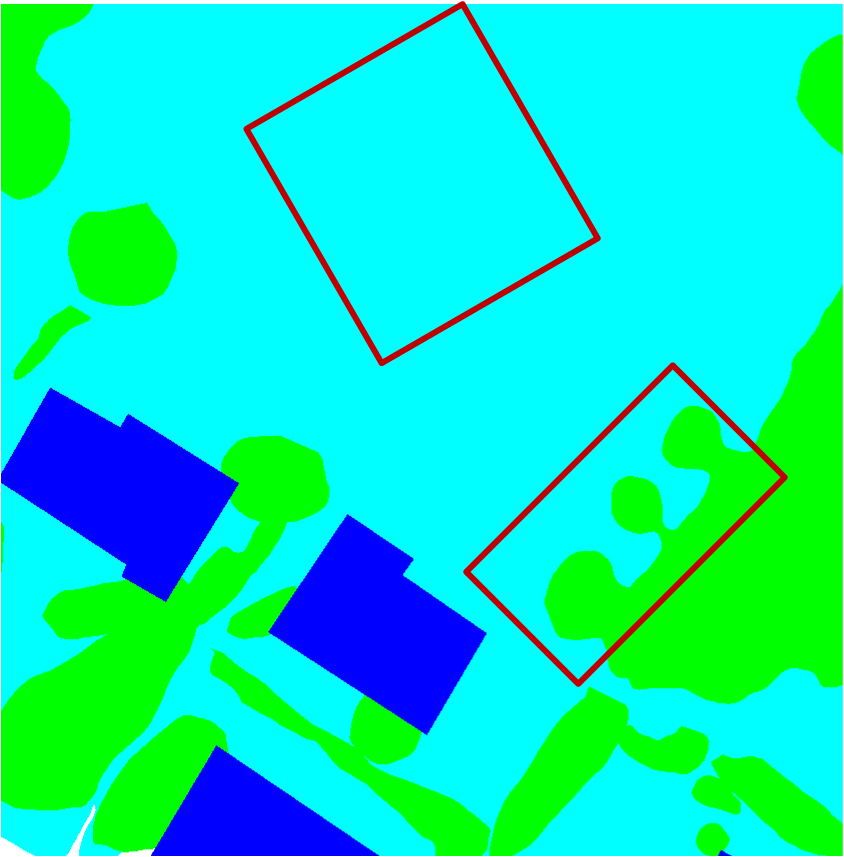}}
		{\includegraphics[width=.1\linewidth]{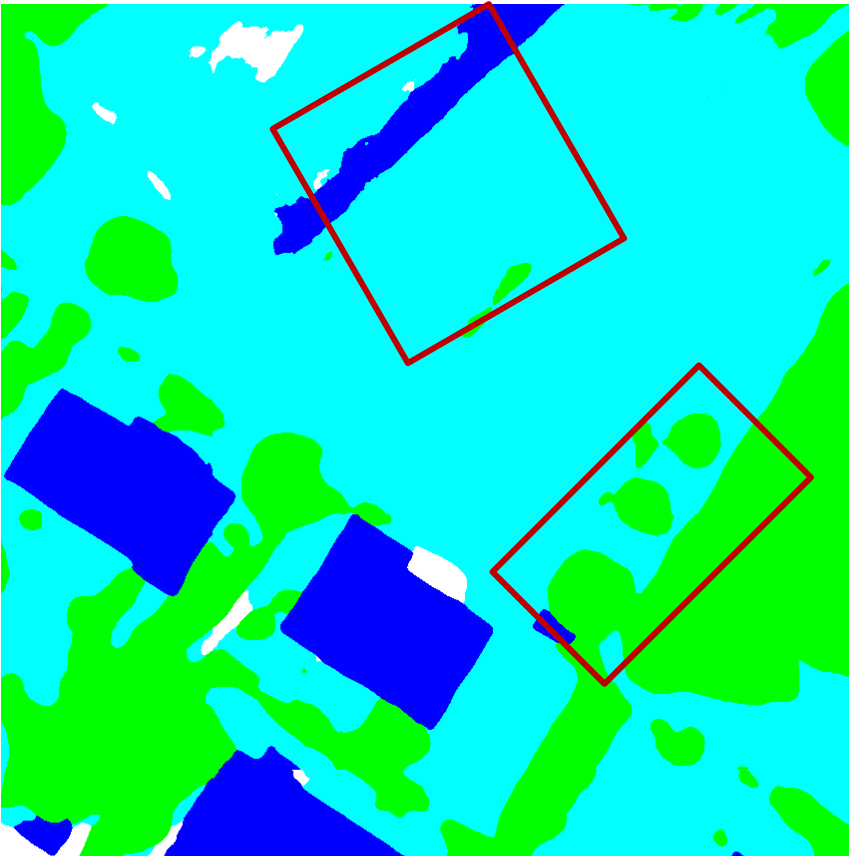}}
		{\includegraphics[width=.1\linewidth]{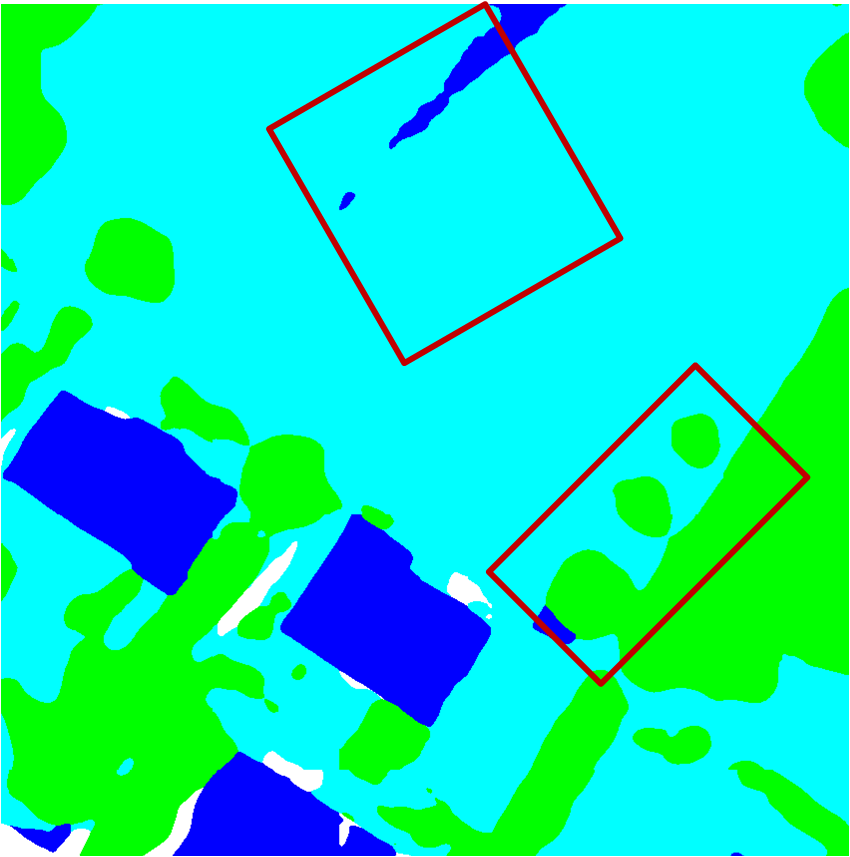}}
		{\includegraphics[width=.1\linewidth]{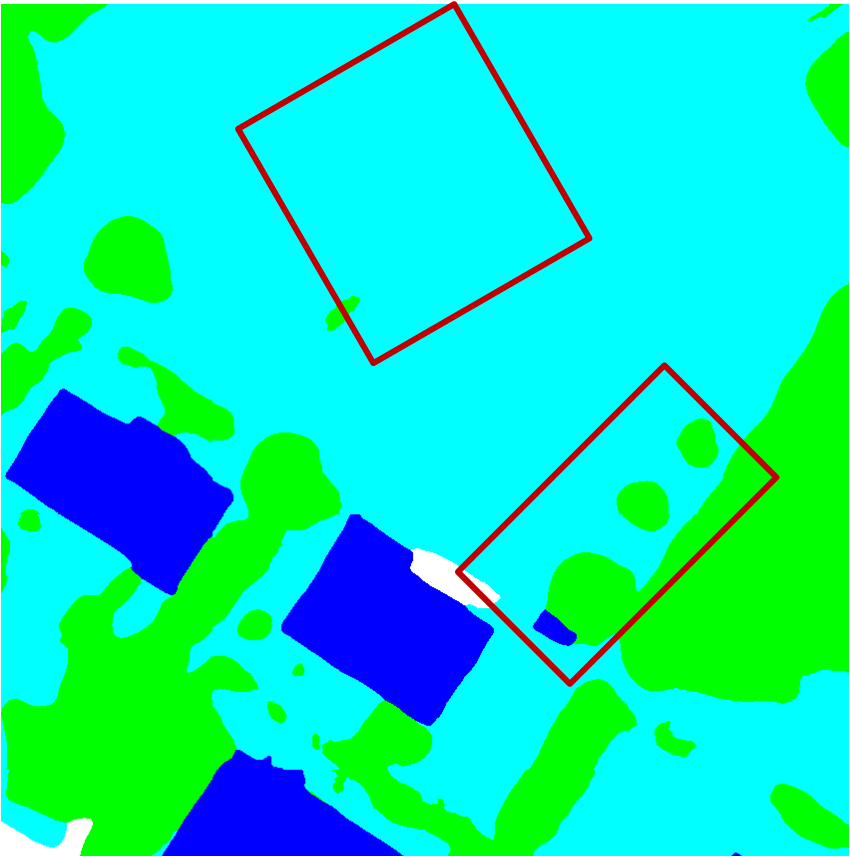}}
		{\includegraphics[width=.1\linewidth]{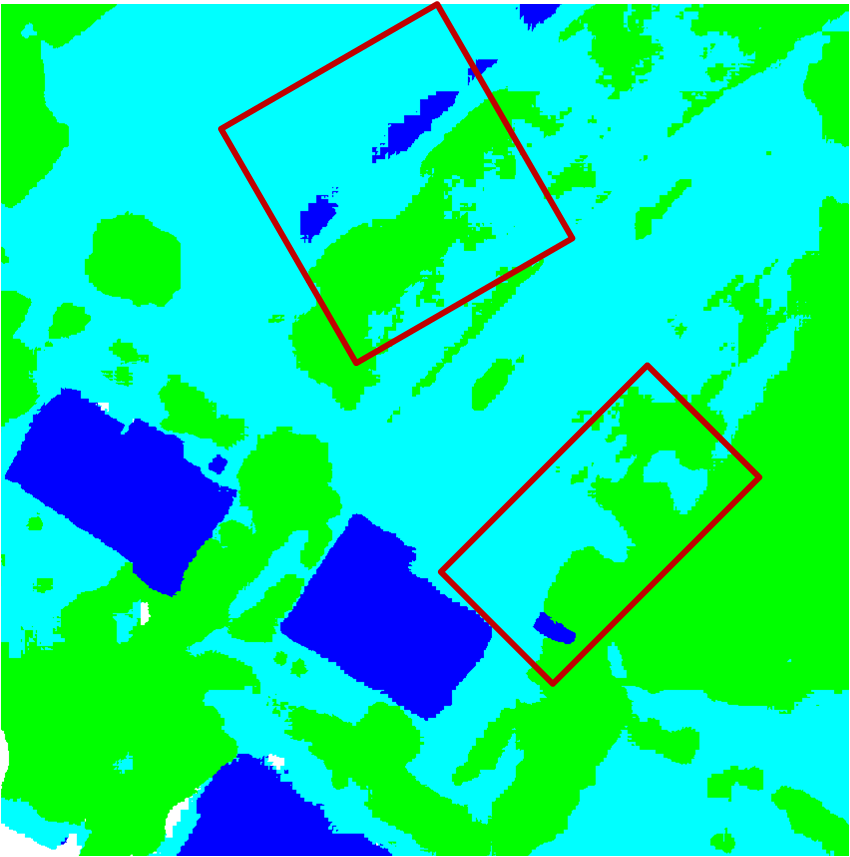}}
		{\includegraphics[width=.1\linewidth]{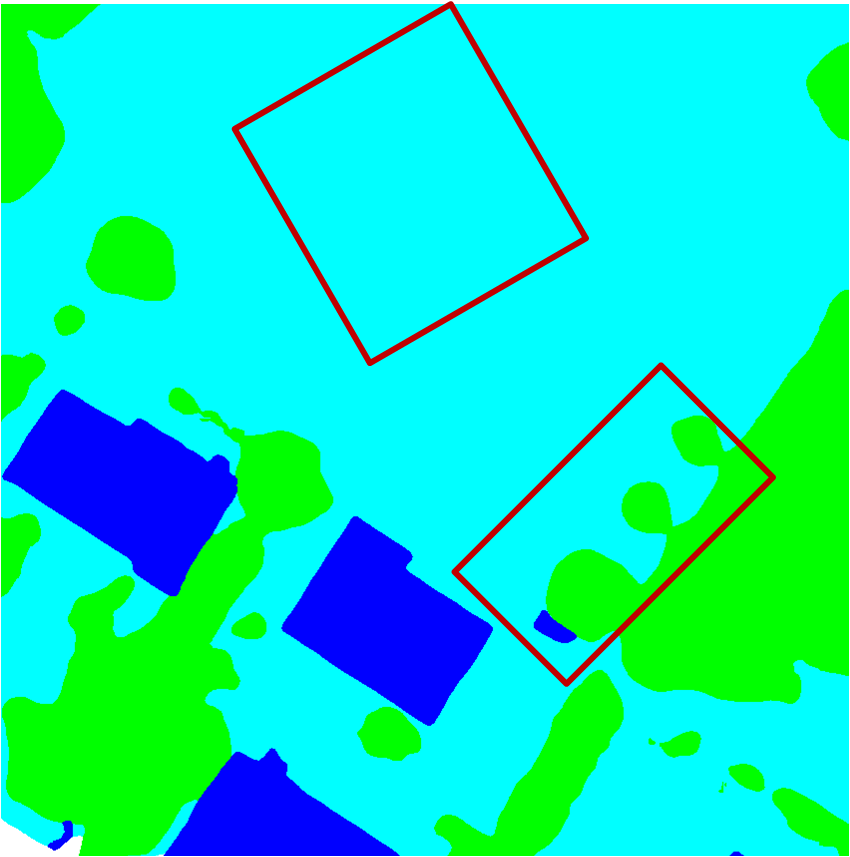}}
		{\includegraphics[width=.1\linewidth]{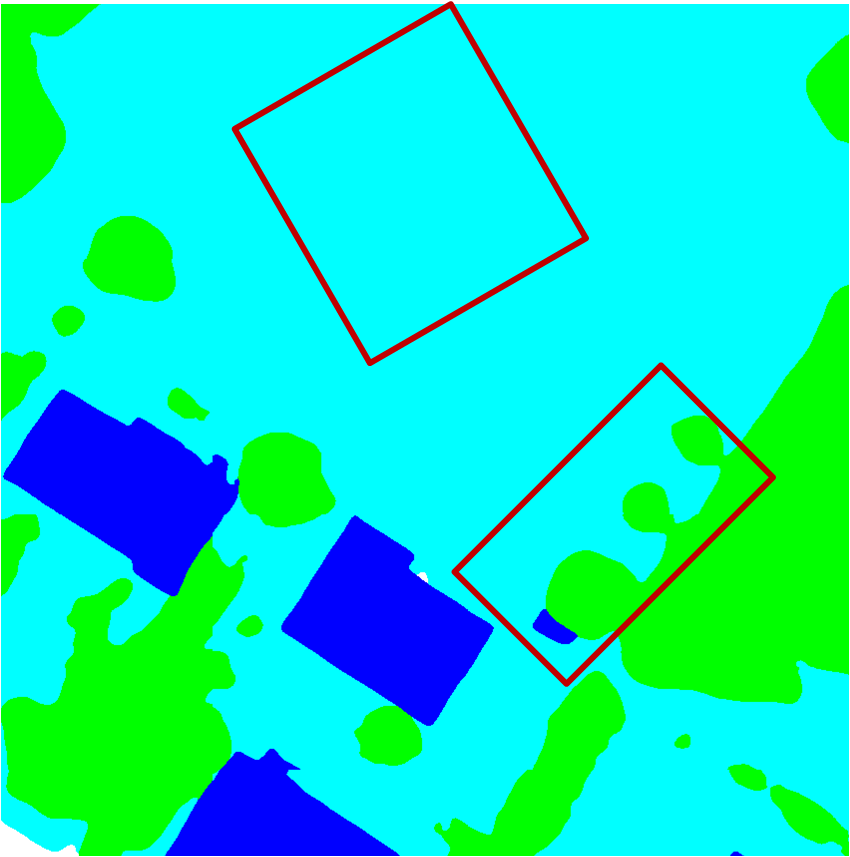}}
		{\includegraphics[width=.1\linewidth]{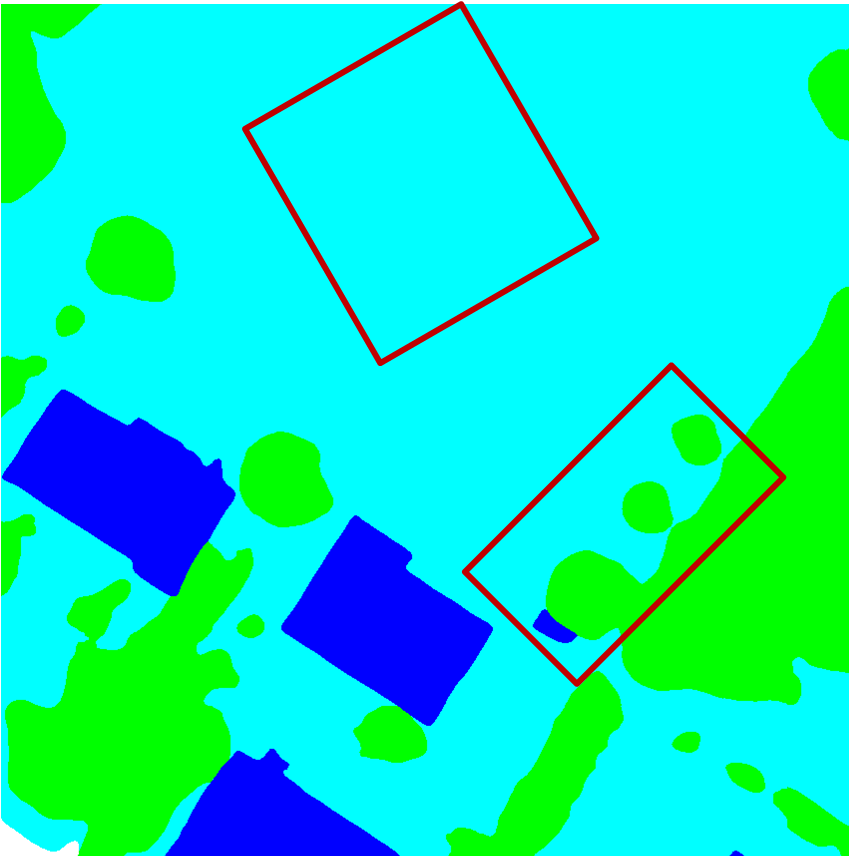}}
	\end{minipage}
	
	\vspace{0.1cm}
	
	\begin{minipage}{\linewidth}
		\centering
		{\includegraphics[width=.1\linewidth]{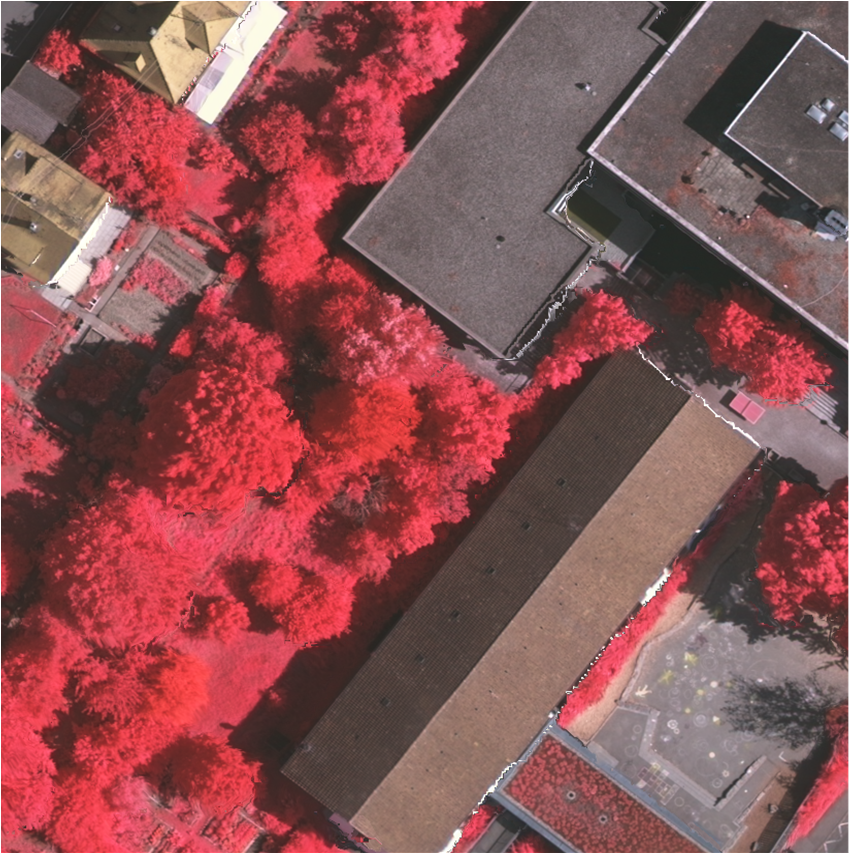}}
		{\includegraphics[width=.1\linewidth]{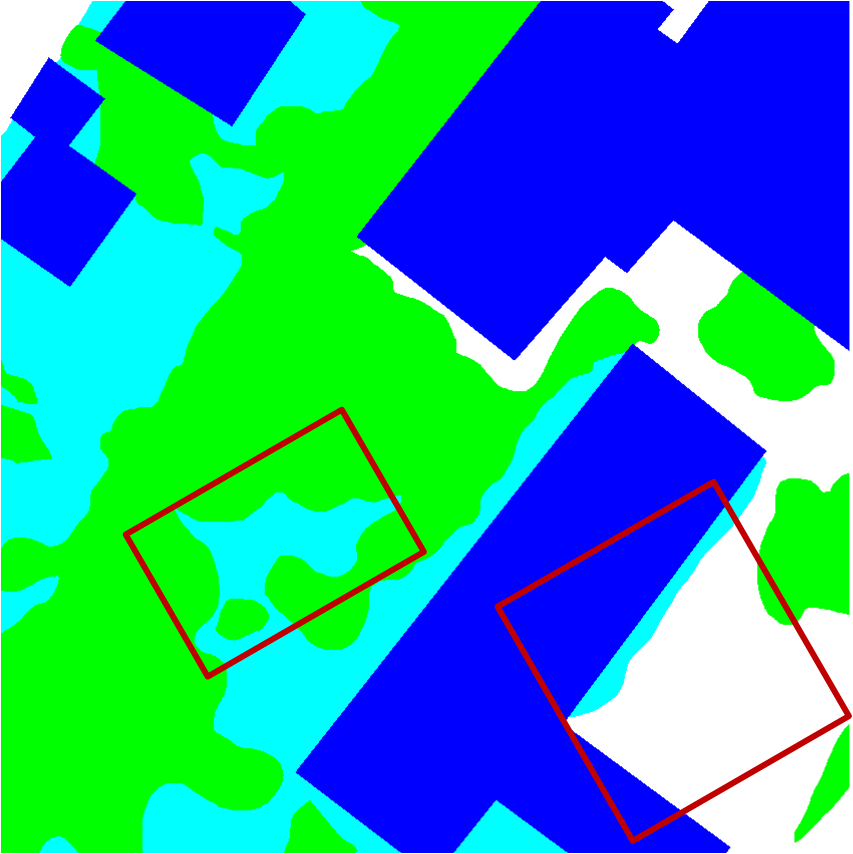}}
		{\includegraphics[width=.1\linewidth]{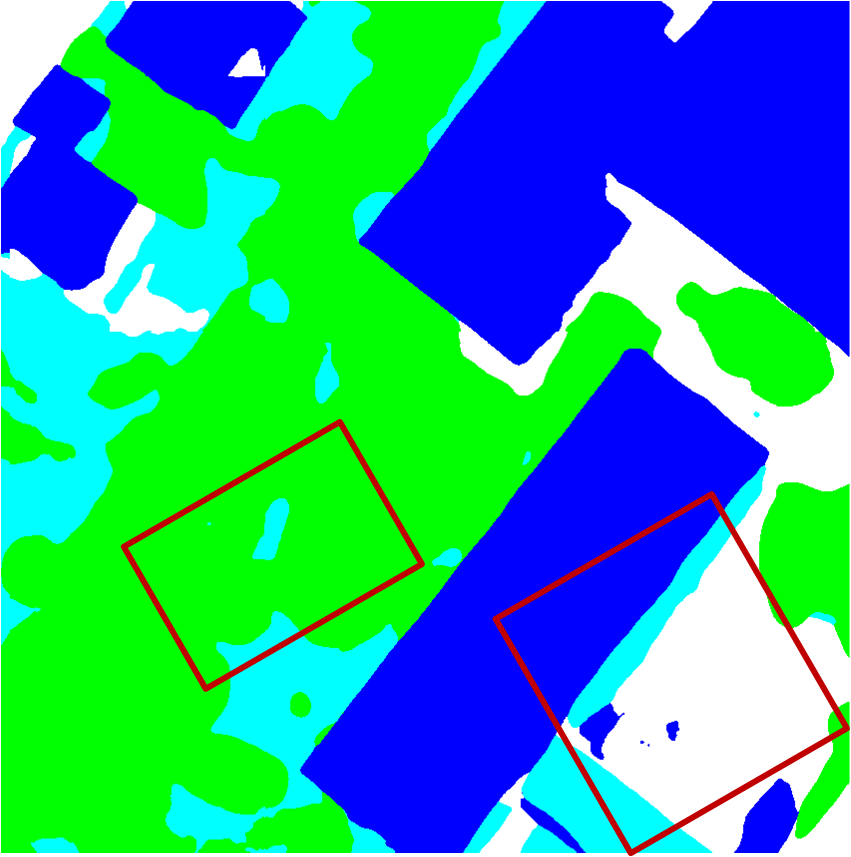}}
		{\includegraphics[width=.1\linewidth]{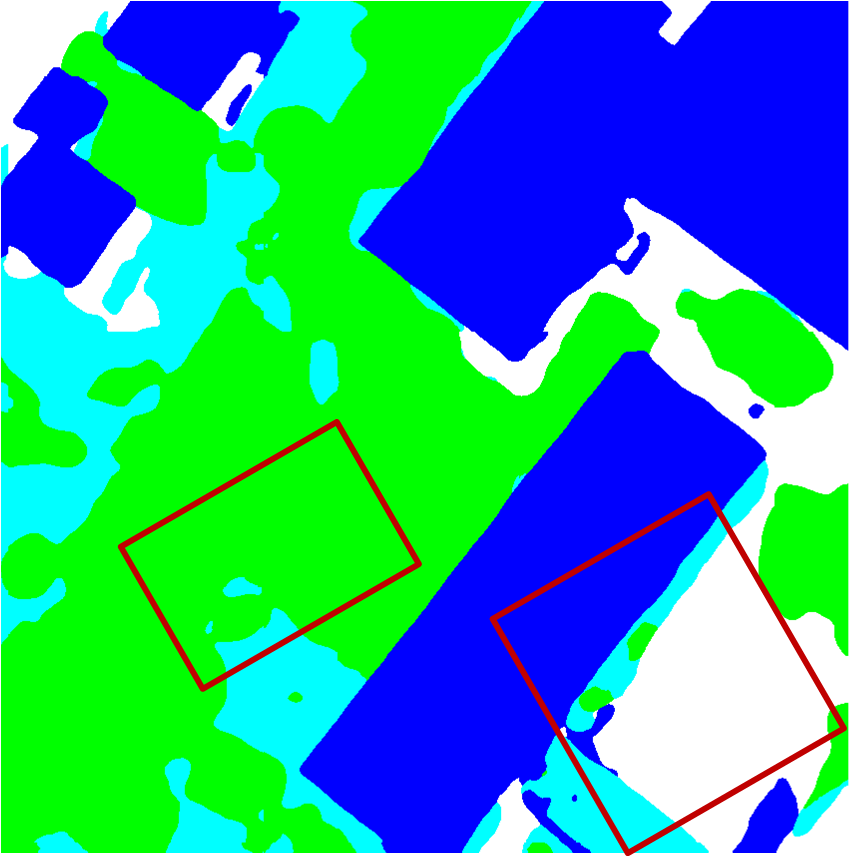}}
		{\includegraphics[width=.1\linewidth]{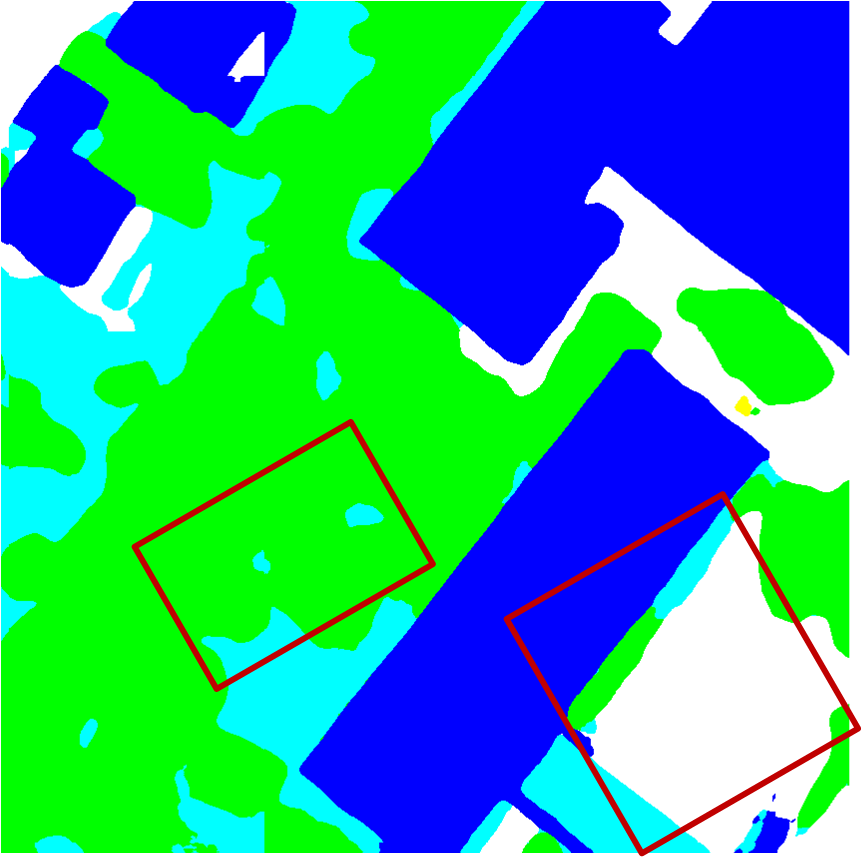}}
		{\includegraphics[width=.1\linewidth]{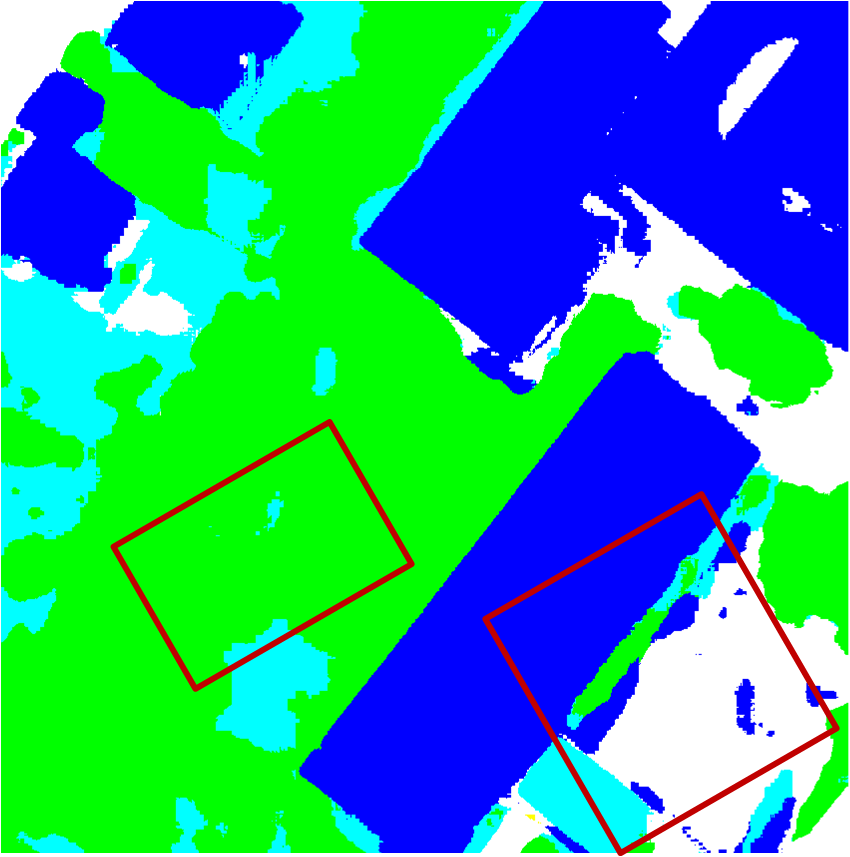}}
		{\includegraphics[width=.1\linewidth]{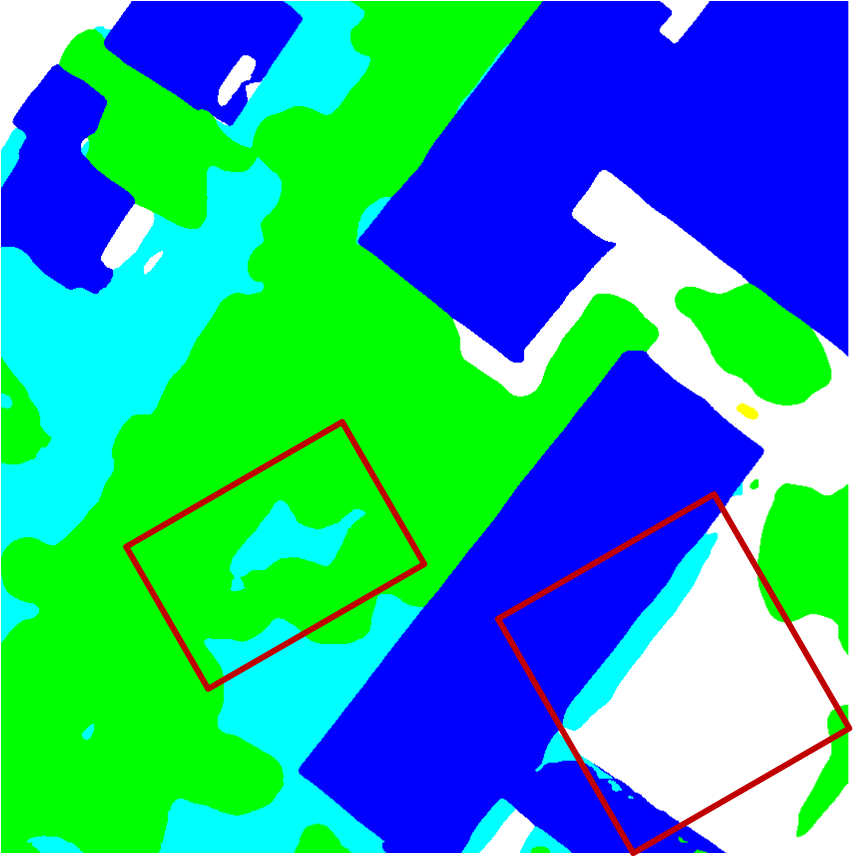}}
		{\includegraphics[width=.1\linewidth]{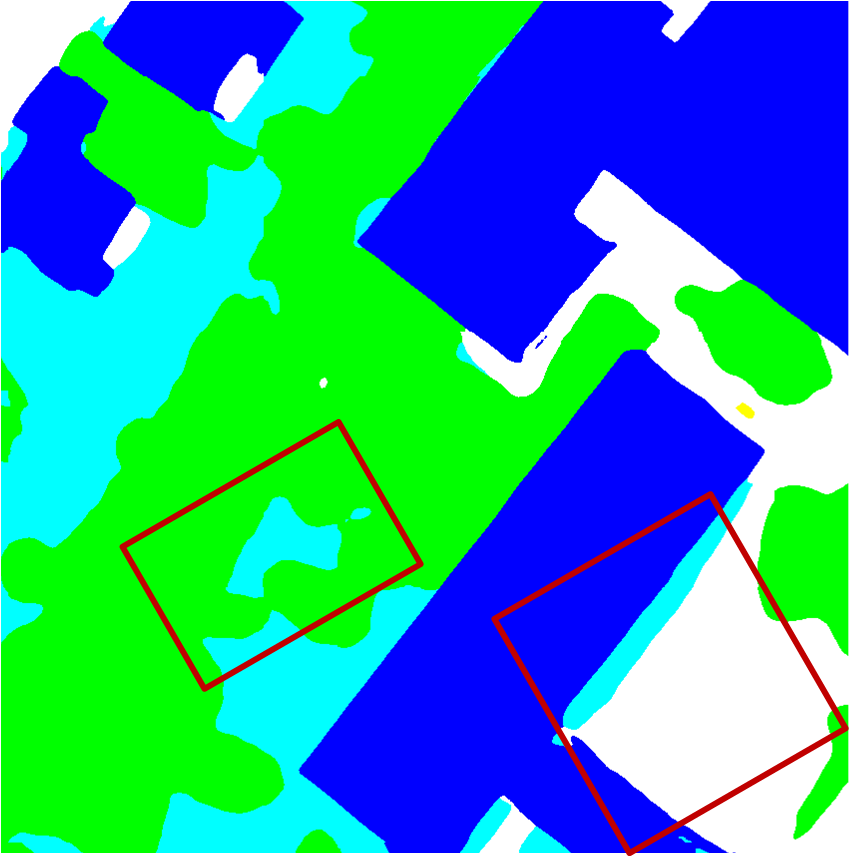}}
		{\includegraphics[width=.1\linewidth]{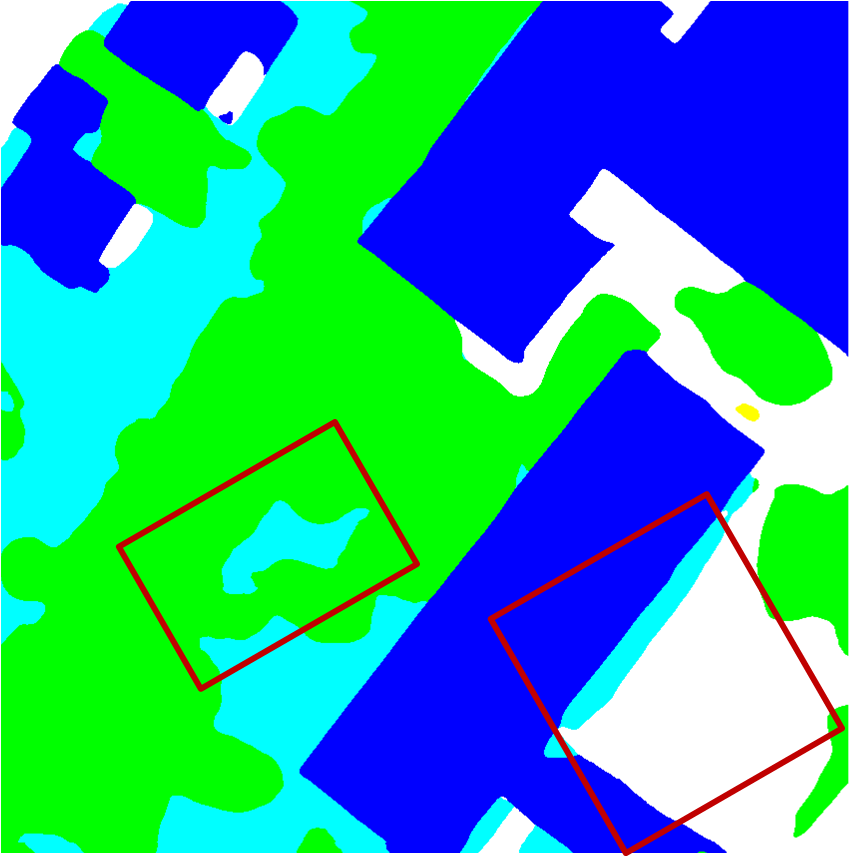}}
	\end{minipage}
	
	\begin{minipage}{\linewidth}
		\setlength\tabcolsep{1.8pt}
		\begin{tabular}{p{.1\linewidth}<\centering p{.1\linewidth}<\centering p{.1\linewidth}<\centering p{.1\linewidth}<\centering p{.1\linewidth}<\centering p{.1\linewidth}<\centering p{.1\linewidth}<\centering p{.1\linewidth}<\centering p{.1\linewidth}<\centering}
			\qquad	(a) & \qquad (b)  &  \;  \quad (c) &  \quad (d)&   \;  (e)  &  \quad (f) &   \quad (g) &  \quad (h) &  \qquad (I)
		\end{tabular}
	\end{minipage}
	\caption{Some samples of ablation experiments. (a) Image. (a) Label. (c) Unet. (d) ResUnet (e) Attention Unet. (f) Swin Unet. (g) GTNet. (h) GTNet + GAN. (I) GATrans.}
	\label{fig7}
\end{figure*}

\subsection{Ablation Experiment}

As shown in Table \ref{tab1}, we conducted a comprehensive evaluation of various methods on the Vaihingen test set. The GTNet outperformed classical networks such as Unet, ResUnet50, Attention Unet, and Swin Unet in terms of both OA and mean F1 score. Furthermore, by incorporating the proposed GAN strategy and structural similarity loss, the performance of GTNet was further enhanced. We present some sample results from our ablation experiments in Figure \ref{fig7}, where areas where the GATrans framework outperforms other methods are highlighted with red boxes. Compared to other ablation methods, the GATrans model demonstrates more precise object predictions with detailed features and smoother boundaries.

\subsection{Comparison Experiment}

\begin{figure}[htb]
\centering
 \includegraphics[width=\linewidth]{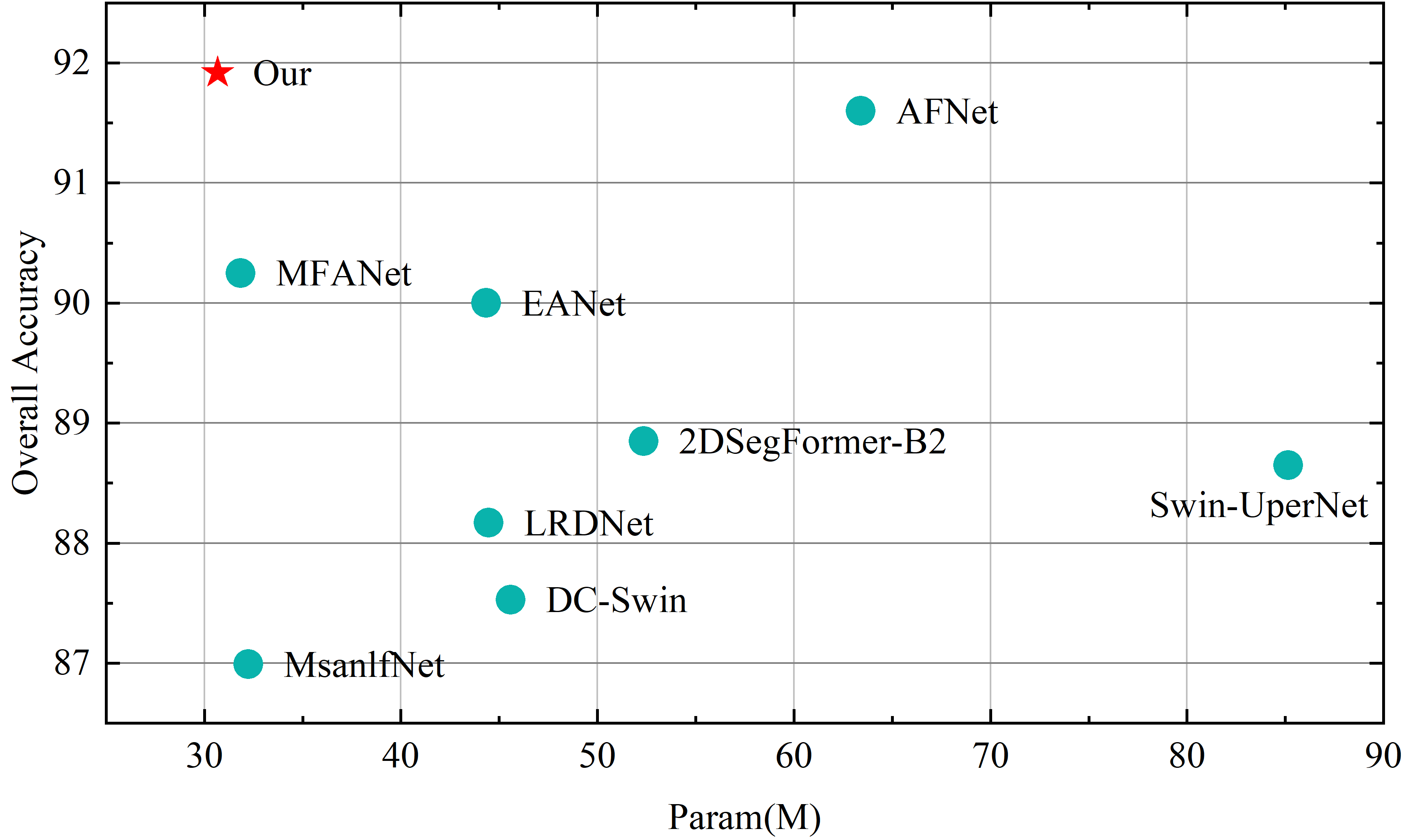}
	\caption{GATrans vs. latest segmentation networks for remote sensing (Parameters and Accuracy). \label{fig1}}
\end{figure}

\begin{table*}[htb]
	\caption{Quantified results of comparison experiments on the test set. \label{tab5}}
        \centering
		\begin{tabular}{lcccccccc}
			\toprule
			\multirow{2.3}{*}{Method} & \multicolumn{5}{c}{F1 Score (\%)} & \multirow{2.3}{*}{OA} &  \multirow{2.3}{*}{\begin{tabular}[c]{@{}c@{}}Mean F1\end{tabular}} &  \multirow{2.3}{*}{{ Param }} \\ 
	
			\cmidrule(lr){2-6} &{Imp surf} & {Building} & {Low veg} & {Tree}   & {Car}   &                                   &         &                   \\ \midrule
            MsanlfNet \cite{12}     & 89.54  & 93.36  & 75.89  & 85.26  & 72.04  & 86.99  & 83.22  & 32.24M  \\
            DC-Swin \cite{13}     & 89.37  & 92.65  & 81.02  & 85.58  & 75.29  & 87.53  & 84.78  & 45.58M \\
            LRDNet \cite{14}     & 91.32  & 93.16  & 80.1   & 87.27  & 74.56  & 88.17  & 85.28  & 44.47M  \\
            Swin-UperNet \cite{15}  & 90.11  & 93.64  & 82.36  & 87.28  & 77.55  & 88.65  & 86.19  & 85.14M \\
            2DSegFormer-B2 \cite{16} & 90.96  & 94.5   & 81.44  & 87.2   & 81.29  & 88.85  & 87.08  & 52.35M \\
            EANet \cite{17}     & 92.17 & 95.20 & 82.81 & 89.25 & 80.56 & 89.99 & 87.99 & 44.34M\\
            MFANet \cite{18}     & 92.55  & 95.27  & 83.86  & 89.12  & 84.78  & 90.25  & 89.12  & 31.85M \\
            GloReNet  \cite{19}     & 92.90   & 95.80   & 84.70   & 90.10   & 86.50   & 91.10   & 90.00      & —   \\
            AFNet  \cite{20}        & 93.40   & 95.90   & 86.00     & 90.70   & 87.20   & 91.60   & 90.64   & 63.40M  \\
            DCFAM  \cite{21}        & 93.60   & 96.18  & 85.75  & 90.36  & 87.64  & 91.63  & 90.71  & —     \\
			Our            & 93.16  & 96.12  & 84.68  & 89.83  & 87.06  & 91.92  & 90.17   & 30.68M \\ \bottomrule
	\end{tabular}
\end{table*}
According to the results presented in Table \ref{tab5}, the comparative experiments between GATrans and other advanced methods demonstrate that GATrans achieves the best performance in VHR image segmentation. GATrans achieves remarkable results with the mean F1 score of 90.17\% and the OA of 91.92\%. It is worth noting that the incorporation of the generative-adversarial strategy and the structural similarity loss in GATrans only affects the training period and does not increase the number of parameters or testing time.

Although there are slight variations in the testing time of GTNet, GTNet+GAN, and GATrans due to random errors in the running device, the differences are negligible. Furthermore, GATrans exhibits efficient performance, with parameters totaling 30.68M and a running time of 1.244 seconds. In comparison to other advanced methods, GATrans emerges as an effective and accurate segmentation technique, making it suitable as an automated remote sensing segmentation tool that can be deployed on mobile devices.

\section{Conclusion} \label{sec6}
We propose an efficient GATrans framework for remote sensing image segmentation by incorporating a generative-adversarial strategy. The framework leverages the efficient GTNet model to capture global features. The GTNet employs multiple GLAM modules, employing the SLH algorithm and the ASS similarity function to categorize global features into distinct query buckets. Our experiments demonstrate the effectiveness of the GATrans framework in remote sensing image segmentation.


\bibliographystyle{IEEEtran}

%
%
%
%
%
%

\vfill
\end{CJK}
\end{document}